\documentclass{svproc}
\usepackage[utf8]{inputenc}

\usepackage{comment}
\usepackage{hyperref}
\usepackage{listings}
\usepackage[dvipsnames]{xcolor}
\usepackage{float}
\usepackage{CJKutf8}
\usepackage{amsmath}
\usepackage{graphicx}

\begin{document}

\title{Comparing BERT against traditional machine learning text classification}
\author{Santiago Gonz\'alez-Carvajal, Eduardo C. Garrido-Merch\'an}

\institute{Universidad Polit\'ecnica de Madrid, Madrid, Spain \and 
Universidad Aut\'onoma de Madrid, Madrid, Spain\\
\email{santiago.gonzalez-carvajal@alumnos.upm.es} \\ 
\email{eduardo.garrido@uam.es}}

\maketitle

\begin{abstract}
The BERT model has arisen as a popular state-of-the-art model in recent years. It is able to cope with NLP tasks such as supervised text classification without human supervision. Its flexibility to cope with any corpus delivering great results has make this approach very popular in academia and industry. Although, other approaches have been used before successfully. We first present BERT and a review on classical NLP approaches. Then, we empirically test with a suite of different scenarios the behaviour of BERT against traditional TF-IDF vocabulary fed to machine learning algorithms. The purpose of this work is adding empirical evidence to support the use of BERT as a default on NLP tasks. Experiments show the superiority of BERT and its independence of features of the NLP problem such as the language of the text adding empirical evidence to use BERT as a default technique in NLP problems. 
\end{abstract}

\section{Introduction}
Natural Language Processing (NLP) methodologies have flourished and lots of papers solving different tasks of the field, such as text classification \cite{aggarwal2012survey}, named entity recognition \cite{nadeau2007survey} or summarization \cite{puente2013creating}, have been published. We can differentiate, mainly, between two types of approaches to NLP problems: Firstly, linguistic approaches \cite{cambria2014jumping} that generally use different features of the text that the experts on the domain consider that are relevant have been extensively used. Those features could be combinations of words, or n-grams \cite{stamatatos2011plagiarism}, grammatical categories, unambiguous meanings of words, words appearing in a particular position, categories of words and much more. These features could be built manually for an specific problem or can be retrieved by using different linguistic resources \cite{besanccon2010lima} such as ontologies \cite{busch2006ontology}.

On the other hand, Machine Learning (ML) \cite{manning1999foundations} and deep learning based approaches \cite{otter2020survey} that classically have analyzed annotated corpora of texts inferring which features of the text, typically in a bag of words fashion \cite{zhang2010understanding} or by n-grams, are relevant for the classification automatically. Both approaches have their pros and cons, concretely, linguistic approaches have great precision but their recall is low as the context where the features are useful is not as big as the one processed by machine learning algorithms. Although, the precision of classical NLP systems was, until recently, generally better as the one delivered by machine learning \cite{garridoexpansion}. Nevertheless, recently, thanks to the rise of computation, machine learning text classification dominates in scenarios where huge sizes of texts are processed.

Generally, linguistic approaches consist in applying a series of rules, which are designed by linguistic experts \cite{khurana2017natural}. An example of linguistic approach can be found at \cite{hutto2014vader}. The advantage of these type of approaches over ML based approaches is that they do not need large amounts of data. Regarding ML based approaches, they usually have a statistical base \cite{khurana2017natural}. We can find many examples of these type of approaches: BERT \cite{devlin2018bert}, Transformers \cite{vaswani2017attention}, etc.\\

Another issue with traditional NLP approaches is multilingualism \cite{bikel2012multilingual}. We can design rules for a given language, but sentence structure, and even the alphabet, may change from one language to another, resulting in the need to design new rules. Some approaches such as the Universal Networking Language (UNL) standard \cite{uchida2001universal} try to circumvent this issue, but the multilingual resource is hard to build and requires experts on the platform. Another problem with UNL approaches and related ones, would be that, given a specific language, the different forms of expression, i.e. the way we write in, for example, Twitter, is very different from the way we write a more formal document, such as a research paper \cite{farzindar2015natural}.

Bidirectional Encoder Representations from Transformers (BERT) is a NLP model that was designed to pretrain deep bidirectional representations from unlabeled text and, after that, be fine-tuned using labeled text for different NLP tasks \cite{devlin2018bert}. That way, with BERT model, we can create state-of-the-art models for many different NLP tasks \cite{devlin2018bert}. We can see the results obtained by BERT in different NLP tasks at \cite{devlin2018bert}.

In this work we compare BERT model \cite{devlin2018bert} with a traditional machine learning NLP approach that trains machine learning algorithms in features retrieved by the Term Frequency - Inverse Document Frequency (TF-IDF) \cite{yun2005improved} algorithm as a representative of these traditional approaches \cite{trstenjak2014knn}. With this technique, we avoid the construction of a linguistic resource that need expert supervision, simulating it with the punctuation retrieved for any term by the TF-IDF technique. We lose precision by doing this operation but gain recall.

We have carried out four different experiments about text classification. In all of them, we have used two different classifiers: BERT and a traditional classifier created in the way that we have just explained. In this work we start by presenting some related work, then, we describe the models we have used in our experiments, after that, we describe the experiments we have carried out and show the obtained results and, finally, we present the conclusions drawn from the work and some future lines of work.

\section{Related Work}
In this section, we summarize the main comparisons against advanced models such as the BERT transformer and classical natural language processing. Recently, BERT has achieved state-of-the-art results in a broad range of NLP tasks \cite{devlin2018bert}, so the question that is discussed is whether classical NLP techniques are still useful in comparison to the outstanding behaviour of BERT and related models.

It is interesting to study how does the BERT model represent the steps of the traditional NLP pipeline \cite{tenney2019bert} in order to make a fair comparison. The main conclusion of this paper is that their work shows that the model adapts to the classical NLP pipeline dynamically, revising lower-level decisions on the basis of disambiguating information from higher-level representations. In other words, we can think of BERT as a generalization of the traditional NLP pipeline, hence being more dynamic.

An argument that defends classical machine learning NLP approaches is that the BERT approach need huge amounts of texts to deliver proper results. An interesting work \cite{usherwood2019low} that focus on a pure empirical comparison of BERT and ULMFiT \cite{rother2018ulmfit} w.r.t traditional NLP approaches in low-shot classification tasks where we only have $100$-$1000$ labelled examples per class shows how BERT, representing the best of deep transfer learning, is the best performing approach, outperforming top classical machine learning algorithms thanks to the use of transfer learning \cite{devlin2018bert}. In our work, we are going to test this hypothesis under different problems that also involve texts in different languages.

A common critique of classical NLP practitioners is that the BERT model and machine learning methodologies can be fooled easily, commiting errors that may be severe in certain applications and that can be easily solved by symbolic approaches. Following this reasoning, in this work \cite{jin2019bert} the authors present the TextFooler baseline, that generates adversarial text in order to fool BERT's classification \cite{jin2019bert}. We wonder if these experiments are representative of common scenarios and hypothesize that, although it is true that some texts may fool BERT, they are not representatives of common problems. In order to test this hypothesis, we are going to measure the results given by BERT in different languages. If BERT fail in these problems, then these adversaries may be common. Although, if BERT outperforms classical approaches under standard circumstances, then we can state that these adversarial attacks may not be common.

\section{The BERT model and the traditional machine learning NLP methodology}\label{section:models}
Having reviewed related work, we will now introduce the traditional NLP approaches that we are comparing with BERT and then, the details of the BERT model. 

\subsection{Term Frequency - Inverse Document Frequency (TF-IDF)}
A classical way to deal with a supervised learning NLP task is to build a bag-of-words model with the most weighted words given by the TF-IDF algorithm. Assuming there are $N$ documents in the collection, and that term $t_i$ occurs in $n_i$ of these documents. Then, inverse document frequency can be computed as:
\begin{equation}\label{eq-def-tfidf}
    idf(t_i) = log \dfrac{N}{n_i}.
\end{equation}
Actually, the original measure was an integer approximation to this formula, and the logarithm was base $2$. However, \eqref{eq-def-tfidf} is the most commonly cited form of IDF. For more information we refer the reader to the original source \cite{robertson2004understanding}.

On the other hand, given a term $t_i$, we denote by $tf_i$ the frequency of the term $t_i$ in the document under consideration \cite{robertson2004understanding}.

Finally, TF-IDF is defined for a given term $t_i$ in a given document as follows:
\begin{center}
    $tfidf(t_i)=tf_i \cdot idf(t_i)$.
\end{center}
In our experiments, regarding the standard NLP algorithms, we will be using TF-IDF to build a vocabulary for a machine learning model. Further details are introduced in the experiments section. 

\subsection{Bidirectional Encoder Representations from Transformers (BERT)}
We now explain what we consider to be the state-of-the-art technique on natural language processing. Regarding the BERT model, there are two steps in its framework: \emph{pre-training} and \emph{fine-tuning} \cite{devlin2018bert}. During pre-training, the model is trained on unlabeled large corpus. For fine-tuning, the model is initialized with the pre-trained parameters, and all the parameters are fine-tuned using labeled data for specific tasks.

BERT's model architecture is a multi-layer bidirectional Transformer encoder \cite{devlin2018bert} based on the original implementation described in \cite{vaswani2017attention}.

This kind of encoder is composed of a stack of $N=6$ identical layers. Each of these layers has two sub-layers. The first one is a multi-head self-attention mechanism, and the second one, is a simple position-wise fully connected feed-forward network. It employs a residual connection \cite{he2016deep} around both sub-layers, followed by a layer normalization \cite{ba2016layer}. That is, the output of each sub-layer is $LayerNorm(x + Sublayer(x))$, where $Sublayer(x)$ is the function implemented by the sub-layer \cite{vaswani2017attention}.

In relation to, multi-head self-attention, first, we need to define scaled dot-product attention. It is define as follows:
\begin{center}
    $Attention(Q,K,V)=softmax(\dfrac{QK^T}{\sqrt{d_k}})V$,
\end{center}
where $Q$ is the matrix of queries, $K$ is the matrix of keys, $V$ is the matrix of values and $d_k$ is the dimension of the $Q$ and $K$ matrices. Now, we can define multi-head attention as
\begin{center}
    $MultiHead(Q,K,V)= Concat(head_1,...,head_h)W^O$,
\end{center}
where $head_i = Attention(QW_i^Q, KW_i^K, VW_i^V)$. Multi-head attention consists on projecting the queries, keys and values $h$ times with different, learned linear projections to $d_k$, $d_k$ and $d_v$ (dimension of the values matrix), respectively. Then, on each of these projected versions of the queries, keys and values, we perform the attention function in parallel, yielding in $d_v$-dimensional output values. Finally, these are concatenated and projected, resulting in the final values \cite{vaswani2017attention}. Self-attention means that all of the keys, values and queries come from the same place.

BERT represents a single sentence or a pair of sentences (for example, the pair $\langle question, answer\rangle$) as a sequence of tokens according to the following features: BERT uses WordPiece embeddings \cite{wu2016google}. The first token of the sequence is ``[CLS]''. When there is a pair of sentence, in the sequence, they are separated by the ``[SEP]'' token. And, an embedding is added to every token indicating whether it belongs to the first or the second sentence. For a given token, its input representation is constructed by summing the corresponding token, position, and segment embeddings \cite{devlin2018bert}.

\textbf{Pre-training} is divided into: \emph{Masked LM} and \emph{Next Sentence Prediction (NSP)}. The first one, consists in masking some percentage of the input tokens at random (using the ``[MASK]'' token), and then, predict those masked tokens. The second one consists in, given two sentences \texttt{A} and \texttt{B}, 50\% of the time \texttt{B} is the actual next sentence that follows \texttt{A} (labeled as \texttt{IsNext}), and 50\% of the time \texttt{B} is a random sentence from the corpus (labeled as \texttt{NotNext}) \cite{devlin2018bert}.

\textbf{Fine-tuning} is straightforward since the self-attention mechanism in the Transformer allows BERT to model many downstream tasks.
For each task, we simply plug in the specific inputs and outputs into BERT and fine-tune all the parameters \cite{devlin2018bert}.

\section{Experiments}
In order to compare BERT model with respect to the traditional machine learning NLP methodology, we have designed four experiments that are described throughout the section.

In these experiments, we will be using \texttt{TfidfVectorizer} from \texttt{sklearn} Py\-thon 3 module. After using TF-IDF to preprocess the text, we will be using \texttt{Predictor} from \texttt{auto\_ml} module (in the third and fourth experiments), and \texttt{H2OAutoML} from \texttt{h2o} module (in the second experiment), to find the best model to fit the data. In the first experiment, we will, instead, show how much work needs to be done in order to get close to the results obtained, with no effort, using BERT model. For this purpose, we will be using many \texttt{sklearn} models and study their results in depth.

Regarding BERT's implementation, we have used the pre-trained BERT model from \texttt{ktrain} Python 3 module. This model expects the following directory structure: a directory which must contain two subdirectories: \textbf{train} and \textbf{test}. Each one of them, in turn, must contain one subdirectory per class (named after the name of the class they represent). And, finally, each class directory, must contain the \texttt{`.txt'} files (their name is irrelevant) with the texts that belong to the class they represent.

\subsection{IMDB experiment}

In the first experiment, we have downloaded the IMDB dataset from the following \href{https://ai.stanford.edu/~amaas/data/sentiment/}{website}. It contains 50000 movie reviews (25000 to train the model and 25000 to test it) to perform sentiment analysis, a popular supervised learning text classification task. The dataset is classified into two different classes: Positive and negative movie reviews.

We have compared the behaviour of a pre-trained default BERT model w.r.t different popular machine learning models such as SVC or Logistic Regression that use a vocabulary extracted from a TF-IDF model obtaining the following results:
    \begin{table}[H]
    \begin{center}
    \begin{tabular}{|p{4cm}|p{3.5cm}|}
    \hline
    Model & Accuracy \\
    \hline \hline
    \textbf{BERT} & \textbf{0.9387} \\ \hline
    Voting Classifier & 0.9007\\ \hline
    Logistic Regression & 0.8949 \\ \hline
    Linear SVC & 0.8989 \\ \hline
    Multinomial NB & 0.8771 \\ \hline
    Ridge Classifier & 0.8990 \\ \hline
    Passive Aggresive Classifier & 0.8931 \\ \hline
    \end{tabular}
    \caption{Accuracy retrieved by the different methodologies in the IMDB experiment over the validation set.}
    \label{table:results-IMDB}
    \end{center}
    \end{table}
As we can see, BERT outperforms the rest of the models. It is noteworthy that obtaining these results with the traditional approaches has been far more complicated than obtaining this result with BERT.
    
\subsection{RealOrNot tweets experiment}    
Our second experiment deals with the RealOrNot tweets written in English. We have downloaded the dataset from the following  \href{https://www.kaggle.com/c/nlp-getting-started/data}{website}. The task to solve here is pure binary text classification. It contains tweets classified into two different classes: Tweets about a real disaster and tweets which are not about a real disaster.

We have just used the \textit{tweet} and \textit{class} columns. We have also used the \texttt{re} Python 3 module to preprocess the tweets (\#anything $->$ hashtag, @anyone $->$ entity, etc.). After that, we have generated the directory structure that we need to use BERT model (using 75\% data to train and 25\% data to validate). The obtained results have been summarized in the following table:
    \begin{table}[H]
    \begin{center}
    \begin{tabular}{|p{4cm}|p{3.5cm}|p{3.5cm}|}
    \hline
    Model & Accuracy & Kaggle Score\\
    \hline \hline
    \textbf{BERT} & \textbf{0.8361} & \textbf{0.83640}\\ \hline
    H2OAutoML & 0.7875 & 0.77607\\ \hline
    \end{tabular}
    \caption{RealOrNot experiment results.}
    \label{table:results-ROrN}
    \end{center}
    \end{table}
Finally, we have classified the data from the Kaggle competition with BERT. We have scored \textbf{0.83640}. We can see this result \href{https://www.kaggle.com/c/nlp-getting-started/leaderboard}{here} (Santiago Gonz\'alez). Regarding the traditional approaches, the best classifier from the \texttt{h2o} module has turned out to be the H2OStackedEnsembleEstimator :  Stacked Ensemble with model key StackedEnsemble\_BestOfFamily\_AutoML\_20200221\_120302. And, its score in the competition has been \textbf{0.77607}.
    
\subsection{Portuguese news experiment}
\subsubsection{Description}
Having seen that BERT has outperformed an AutoML technique and other classical machine learning algorithms using a vocabulary built from a traditional NLP technique such as TF-IDF in the English language, we choose to change the language to see if the BERT model also behaves well. We have downloaded the Portuguese news dataset from the following  \href{https://www.kaggle.com/c/fasam-nlp-competition-turma-4/data}{website}. It contains articles from the news classified into nine different classes: ambiente, equilibrioesaude, sobretudo, educacao, ciencia, tec, turismo, empreendedorsocial and comida.

We have just used the \textit{article text} and \textit{class} columns. We have generated the directory structure that we need to use BERT model (using 75\% data to train and 25\% data to validate obtaining the following results:
    \begin{table}[H]
    \begin{center}
    \begin{tabular}{|p{4cm}|p{3.5cm}|p{3.5cm}|}
    \hline
    Model & Accuracy & Kaggle Score \\
    \hline \hline
    \textbf{BERT} & \textbf{0.9093} & \textbf{0.91196}\\ \hline
    Predictor (auto\_ml) & 0.8480 & 0.85047\\ \hline
    \end{tabular}
    \caption{Portuguese news experiment results.}
    \label{table:results-Port}
    \end{center}
    \end{table}
Finally, we have classified the data for the Kaggle competition scoring a \textbf{0.91196} accuracy. We can see this result \href{https://www.kaggle.com/c/fasam-nlp-competition-turma-4/leaderboard}{here} (Santiago Gonz\'alez). Regarding the traditional methods, the best classifier has turned out to be a \texttt{GradientBoostingClassifier}. And, the score in the competition of this model has benn \textbf{0.85047}.
    
\subsection{Chinese hotel reviews experiment}\label{exp-4}
\subsubsection{Description}
Our last experiment involves a completely different language, Peninsular Chinese simplified characters zh-CN, where we hypothesize that, given that the way of expressing this Language is through different symbols that are not separated by spaces BERT may not output a good result. The experiment is a sentiment analysis problem involving Chinese hotel reviews. We have downloaded the dataset from the following  \href{https://github.com/Tony607/Chinese_sentiment_analysis/tree/master/data/ChnSentiCorp_htl_ba_6000}{website}. It contains hotel reviews classified into two different classes: Positive hotel reviews and negative hotel reviews.

In this experiment, we have used 85\% of the data to train the model and 15\% of the data to validate it. Results are given in the following table:
    \begin{table}[H]
    \begin{center}
    \begin{tabular}{|p{4cm}|p{3.5cm}|}
    \hline
    Model & Accuracy\\
    \hline \hline
    \textbf{BERT} & \textbf{0.9381}\\ \hline
    Predictor (auto\_ml) & 0.7399\\ \hline
    \end{tabular}
    \caption{Chinese hotel reviews results.}
    \label{table:results-Chin}
    \end{center}
    \end{table}
We can observe how, independently of the language and its characteristics, BERT behaviour outperforms classical NLP approach.

Finally, we have tried to do some predictions with BERT using Google Translator. For example, we have tried to predict a class for:
\begin{CJK*}{UTF8}{gbsn}
这家酒店的风景和服务都非常糟糕
\end{CJK*}
, which means: "the view and service of this hotel are very bad". The predicted class for this hotel review has been \textbf{neg}, which is correct.
    
Regarding the traditional approaches, the best model has turned out to be a \texttt{GradientBoostingClassifier}. But in this case, the model has been pretty bad, since the probability for both classes is very close. In this experiment, the importance of transfer learning has become apparent, since the dataset was pretty small compared to the ones used in the previous experiments. 

\section{Conclusions and further work}
In this work we have introduced the BERT model and the classical NLP strategy where a machine learning model is trained using the features retrieved with TF-IDF and hypothesize about the behaviour of BERT w.r.t these techniques in the search of a default technique to tackle NLP tasks. We have introduced four different NLP scenarios where we have shown how BERT has outperformed the traditional NLP approach, adding empirical evidence of its superiority in average NLP problems w.r.t. classical methodologies. Furthermore, and of critical interest, implementing BERT has turned out to be far less complicated than implementing the traditional methods. It is also noteworthy the importance of transfer learning. We have been able to obtain this results thanks to pre-training. Transfer learning has become more apparent in experiment \ref{exp-4} (which has the smallest dataset among all the experiments). We are nevertheless aware of the limitations of the BERT model. Although it seems that it is a good default for NLP tasks, its results can be improved. In order to do so, we would like to research in a hyperparameter auto-tuned BERT model for any new NLP task with Bayesian Optimization. We would like to use that auto-tuned BERT to enable classification of language messages for robots \cite{merchan2020machine} \cite{garrido2020artificial} showing consciousness correlated behaviours. 

\section*{Acknowledgments}

The authors gratefully acknowledge the use of the facilities of Centro
de Computaci\'on Cient\'ifica (CCC) at Universidad Aut\'onoma de
Madrid. The authors also acknowledge financial support from Spanish
Plan Nacional I+D+i, grants TIN2016-76406-P and TEC2016-81900-REDT.

\bibliographystyle{plain}
\bibliography{main}

\begin{thebibliography}{10}

\bibitem{aggarwal2012survey}
Charu~C Aggarwal and ChengXiang Zhai.
\newblock A survey of text classification algorithms.
\newblock In {\em Mining text data}, pages 163--222. Springer, 2012.

\bibitem{ba2016layer}
Jimmy~Lei Ba, Jamie~Ryan Kiros, and Geoffrey~E Hinton.
\newblock Layer normalization.
\newblock {\em arXiv preprint arXiv:1607.06450}, 2016.

\bibitem{besanccon2010lima}
Romaric Besan{\c{c}}on, Ga{\"e}l De~Chalendar, Olivier Ferret, Fai{\"\i}za
  Gara, Olivier Mesnard, Meriama La{\"\i}b, and Nasredine Semmar.
\newblock Lima: A multilingual framework for linguistic analysis and linguistic
  resources development and evaluation.
\newblock In {\em LREC}, 2010.

\bibitem{bikel2012multilingual}
Daniel Bikel and Imed Zitouni.
\newblock {\em Multilingual natural language processing applications: from
  theory to practice}.
\newblock IBM Press, 2012.

\bibitem{busch2006ontology}
Justin~Eliot Busch, Albert~Deirchow Lin, Patrick~John Graydon, and Maureen
  Caudill.
\newblock Ontology-based parser for natural language processing, April~11 2006.
\newblock US Patent 7,027,974.

\bibitem{cambria2014jumping}
Erik Cambria and Bebo White.
\newblock Jumping nlp curves: A review of natural language processing research.
\newblock {\em IEEE Computational intelligence magazine}, 9(2):48--57, 2014.

\bibitem{devlin2018bert}
Jacob Devlin, Ming-Wei Chang, Kenton Lee, and Kristina Toutanova.
\newblock Bert: Pre-training of deep bidirectional transformers for language
  understanding.
\newblock {\em arXiv preprint arXiv:1810.04805}, 2018.

\bibitem{farzindar2015natural}
Atefeh Farzindar and Diana Inkpen.
\newblock Natural language processing for social media.
\newblock {\em Synthesis Lectures on Human Language Technologies}, 8(2):1--166,
  2015.

\bibitem{garridoexpansion}
Eduardo~C Garrido and Jes{\'u}s~Cardenosa Lera.
\newblock Expansi{\'o}n supervisada de l{\'e}xicos polarizados adaptable al
  contexto.

\bibitem{garrido2020artificial}
Eduardo~C Garrido-Merch{\'a}n, Martin Molina, and Francisco~M Mendoza.
\newblock An artificial consciousness model and its relations with philosophy
  of mind.
\newblock {\em arXiv preprint arXiv:2011.14475}, 2020.

\bibitem{he2016deep}
Kaiming He, Xiangyu Zhang, Shaoqing Ren, and Jian Sun.
\newblock Deep residual learning for image recognition.
\newblock In {\em Proceedings of the IEEE conference on computer vision and
  pattern recognition}, pages 770--778, 2016.

\bibitem{hutto2014vader}
Clayton~J Hutto and Eric Gilbert.
\newblock Vader: A parsimonious rule-based model for sentiment analysis of
  social media text.
\newblock In {\em Eighth international AAAI conference on weblogs and social
  media}, 2014.

\bibitem{jin2019bert}
Di~Jin, Zhijing Jin, Joey~Tianyi Zhou, and Peter Szolovits.
\newblock Is bert really robust? natural language attack on text classification
  and entailment.
\newblock {\em arXiv preprint arXiv:1907.11932}, 2019.

\bibitem{khurana2017natural}
Diksha Khurana, Aditya Koli, Kiran Khatter, and Sukhdev Singh.
\newblock Natural language processing: State of the art, current trends and
  challenges.
\newblock {\em arXiv preprint arXiv:1708.05148}, 2017.

\bibitem{manning1999foundations}
Christopher~D Manning, Christopher~D Manning, and Hinrich Sch{\"u}tze.
\newblock {\em Foundations of statistical natural language processing}.
\newblock MIT press, 1999.

\bibitem{merchan2020machine}
Eduardo C~Garrido Merch{\'a}n and Mart{\'\i}n Molina.
\newblock A machine consciousness architecture based on deep learning and
  gaussian processes.
\newblock {\em arXiv preprint arXiv:2002.00509}, 2020.

\bibitem{nadeau2007survey}
David Nadeau and Satoshi Sekine.
\newblock A survey of named entity recognition and classification.
\newblock {\em Lingvisticae Investigationes}, 30(1):3--26, 2007.

\bibitem{otter2020survey}
Daniel~W Otter, Julian~R Medina, and Jugal~K Kalita.
\newblock A survey of the usages of deep learning for natural language
  processing.
\newblock {\em IEEE Transactions on Neural Networks and Learning Systems},
  2020.

\bibitem{puente2013creating}
Cristina Puente, Jos{\'e}~Angel Olivas, E~Garrido, and R~Seisdedos.
\newblock Creating a natural language summary from a compressed causal graph.
\newblock In {\em 2013 joint ifsa world congress and nafips annual meeting
  (ifsa/nafips)}, pages 513--518. IEEE, 2013.

\bibitem{robertson2004understanding}
Stephen Robertson.
\newblock Understanding inverse document frequency: on theoretical arguments
  for idf.
\newblock {\em Journal of documentation}, 2004.

\bibitem{rother2018ulmfit}
Kristian Rother and Achim Rettberg.
\newblock Ulmfit at germeval-2018: A deep neural language model for the
  classification of hate speech in german tweets.
\newblock 2018.

\bibitem{stamatatos2011plagiarism}
Efstathios Stamatatos.
\newblock Plagiarism detection using stopword n-grams.
\newblock {\em Journal of the American Society for Information Science and
  Technology}, 62(12):2512--2527, 2011.

\bibitem{tenney2019bert}
Ian Tenney, Dipanjan Das, and Ellie Pavlick.
\newblock Bert rediscovers the classical nlp pipeline.
\newblock {\em arXiv preprint arXiv:1905.05950}, 2019.

\bibitem{trstenjak2014knn}
Bruno Trstenjak, Sasa Mikac, and Dzenana Donko.
\newblock Knn with tf-idf based framework for text categorization.
\newblock {\em Procedia Engineering}, 69:1356--1364, 2014.

\bibitem{uchida2001universal}
Hiroshi Uchida and Meiying Zhu.
\newblock The universal networking language beyond machine translation.
\newblock In {\em International Symposium on Language in Cyberspace, Seoul},
  pages 26--27, 2001.

\bibitem{usherwood2019low}
Peter Usherwood and Steven Smit.
\newblock Low-shot classification: A comparison of classical and deep transfer
  machine learning approaches.
\newblock {\em arXiv preprint arXiv:1907.07543}, 2019.

\bibitem{vaswani2017attention}
Ashish Vaswani, Noam Shazeer, Niki Parmar, Jakob Uszkoreit, Llion Jones,
  Aidan~N Gomez, {\L}ukasz Kaiser, and Illia Polosukhin.
\newblock Attention is all you need.
\newblock In {\em Advances in neural information processing systems}, pages
  5998--6008, 2017.

\bibitem{wu2016google}
Yonghui Wu, Mike Schuster, Zhifeng Chen, Quoc~V Le, Mohammad Norouzi, Wolfgang
  Macherey, Maxim Krikun, Yuan Cao, Qin Gao, Klaus Macherey, et~al.
\newblock Google's neural machine translation system: Bridging the gap between
  human and machine translation.
\newblock {\em arXiv preprint arXiv:1609.08144}, 2016.

\bibitem{yun2005improved}
Zhang Yun-tao, Gong Ling, and Wang Yong-cheng.
\newblock An improved tf-idf approach for text classification.
\newblock {\em Journal of Zhejiang University-Science A}, 6(1):49--55, 2005.

\bibitem{zhang2010understanding}
Yin Zhang, Rong Jin, and Zhi-Hua Zhou.
\newblock Understanding bag-of-words model: a statistical framework.
\newblock {\em International Journal of Machine Learning and Cybernetics},
  1(1-4):43--52, 2010.

\end{thebibliography}

\end{document}